%% file: sam2point.tex
\documentclass{article} 
\usepackage{iclr2024_conference,times}

\input{math_commands.tex}

\usepackage{amsmath}
\usepackage{xspace}

\usepackage{xcolor}         
\usepackage{color, colortbl}
\definecolor{citecolor}{HTML}{2980b9}
\definecolor{linkcolor}{HTML}{c0392b}

\definecolor{darkspringgreen}{rgb}{0.09, 0.45, 0.27}
\definecolor{green(pigment)}{rgb}{0.0, 0.65, 0.31}
\definecolor{skyblue}{rgb}{0.53, 0.81, 0.92}

\usepackage[utf8]{inputenc} 
\usepackage[T1]{fontenc}    
\usepackage{url}            
\usepackage{booktabs}       
\usepackage{amsfonts}       
\usepackage{nicefrac}       
\usepackage{microtype}      
\usepackage{xcolor}         
\usepackage{graphicx}
\usepackage{float}
\usepackage{wrapfig}
\usepackage{makecell}
\usepackage{subfigure}
\usepackage{amsmath}
\usepackage{amssymb}
\usepackage{multirow}
\usepackage{diagbox}
\usepackage{colortbl}
\usepackage{tabularx}
\usepackage{adjustbox}
\usepackage{pifont}
\usepackage{bm}
\usepackage{mathtools}
\usepackage{algpseudocode}
\usepackage{algorithm}
\usepackage{listings}
\usepackage{url}            
\usepackage{booktabs}       
\usepackage{amsfonts}       
\usepackage{nicefrac}       
\usepackage{microtype}      
\usepackage{xcolor}         
\usepackage[hidelinks,breaklinks=true,colorlinks,bookmarks=false,citecolor=citecolor,linkcolor=linkcolor]{hyperref}
\definecolor{DarkGreen}{HTML}{006400}
\definecolor{Darkred}{HTML}{7B241E}
\definecolor{light_red}{HTML}{F7E6E6}
\definecolor{light_green}{HTML}{E9EFEF}
\definecolor{forest_green}{HTML}{228B22}
\definecolor{ggreen}{HTML}{32CD32}
\newcommand{\method}{\textsc{Sam2Point}\xspace}

\title{
\begin{minipage}{.09\textwidth}
\centering
\includegraphics[width=1.2\linewidth]{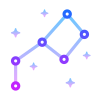}
\end{minipage}%
\begin{minipage}{.9\textwidth}
\begin{center}
\method: Segment Any 3D as Videos\\in Zero-shot and Promptable Manners
\end{center}
\end{minipage}}
    
\author{Ziyu Guo$^{1*}$, Renrui Zhang$^{2,3*\dagger}$, Xiangyang Zhu$^{4*}$, Chengzhuo Tong$^{4}$\\\textbf{Peng Gao$^{4}$, Chunyuan Li$^{3}$, Pheng-Ann Heng$^{1}$}\vspace{0.3cm}\\
  $^1$CUHK MiuLar Lab\quad 
  $^2$CUHK MMLab\quad 
  $^3$ByteDance\quad 
  $^4$Shanghai AI Laboratory\\
\texttt{\{ziyuguo, renruizhang\}@link.cuhk.edu.hk}
\vspace{0.3cm}\\$^*$ Equal contribution \ \ $^{\dagger}$ Project lead
}

\iclrfinalcopy 

\begin{document}
\maketitle

\begin{abstract}
We introduce \textbf{\method}, a preliminary exploration adapting Segment Anything Model 2 (SAM 2) for \textit{zero-shot and promptable} 3D segmentation. 
\method interprets any 3D data as a series of multi-directional videos, and leverages SAM 2 for 3D-space segmentation, without further training or 2D-3D projection. Our framework supports various prompt types, including \textit{3D points, boxes, and masks}, and can generalize across diverse scenarios, such as \textit{3D objects, indoor scenes, outdoor scenes, and raw LiDAR.}
Demonstrations on multiple 3D datasets, e.g., Objaverse, S3DIS, ScanNet, Semantic3D, and KITTI, highlight the robust generalization capabilities of \method. To our best knowledge, we present the most faithful implementation of SAM in 3D, which may serve as a starting point for future research in promptable 3D segmentation.

\vspace{0.3cm}Live Demo: \url{https://huggingface.co/spaces/ZiyuG/SAM2Point}

Code: \url{https://github.com/ZiyuGuo99/SAM2Point}

\end{abstract}
\begin{figure*}[h]
  \centering
\includegraphics[width=\textwidth]{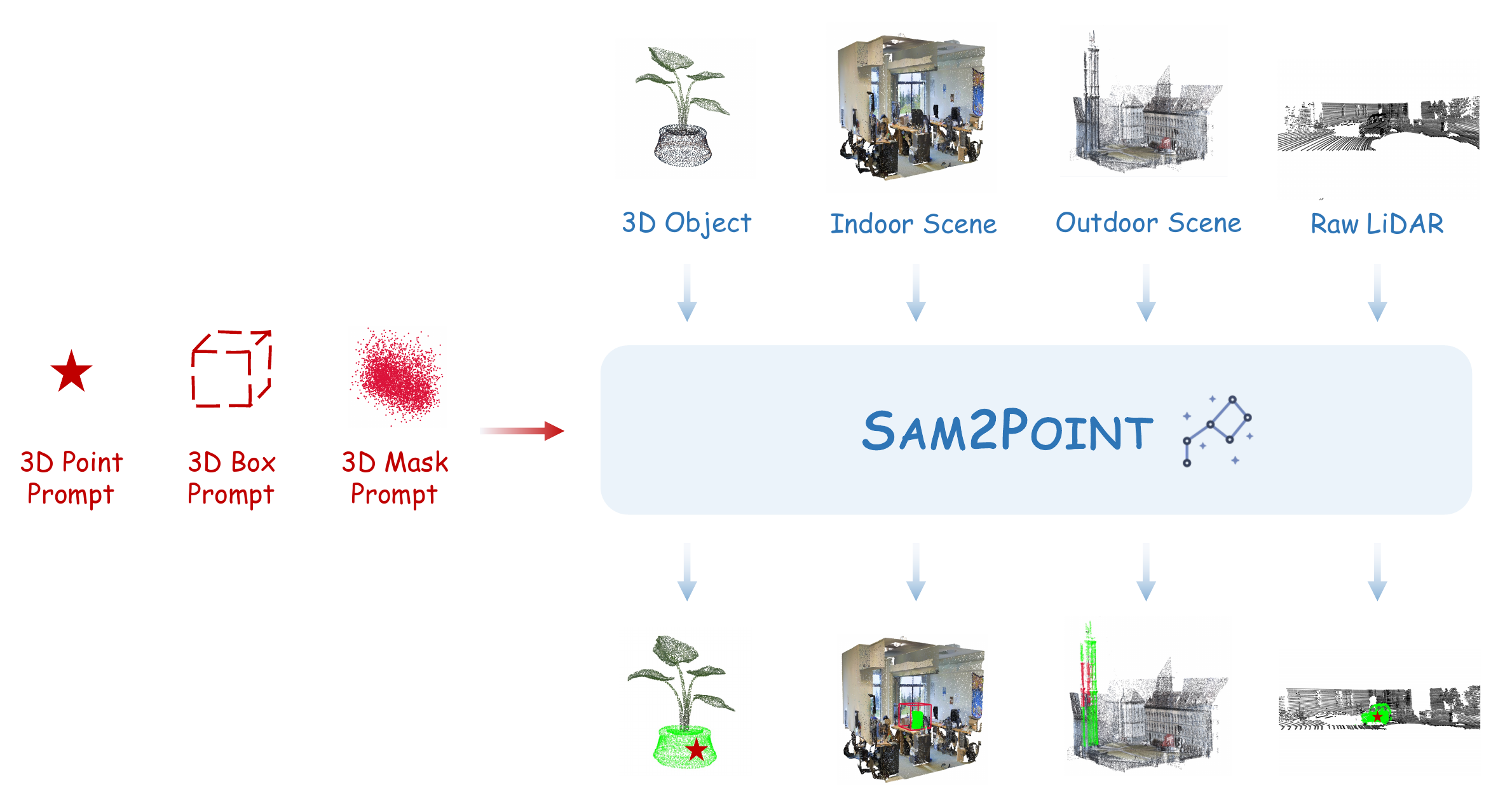}
   \caption{\textbf{The Segmentation Paradigm of \method.} We introduce a zero-shot and promptable framework for robust 3D segmentation via SAM 2~\citep{ravi2024sam}. It supports various user-provided 3D prompt, and can generalize to diverse 3D scenarios. The \textcolor{red}{3D prompt} and \textcolor{ggreen}{segmentation results} are highlighted in red and green, respectively.}
    \label{f1}
\end{figure*}

\section{Introduction}
\label{s1}

Segment Anything Model (SAM)~\citep{kirillov2023segment} has established a superior and fundamental framework for interactive image segmentation. Building on its strong transferability, follow-up research further extends SAM to diverse visual domains, e.g., personalized objects~\citep{zhang2023personalize,liu2023matcher}, medical imaging~\citep{ma2024segment,mazurowski2023segment}, and temporal sequences~\citep{yang2023track,cheng2023segment}. More recently, Segment Anything Model 2 (SAM 2)~\citep{ravi2024sam} is proposed for impressive segmentation capabilities in video scenarios, capturing complex real-world dynamics.

\textit{Despite this, effectively adapting SAM for 3D segmentation still remains an unresolved challenge.}

We identify three primary issues within previous efforts, as compared in Table~\ref{t1}, which prevent them from fully leveraging SAM's advantages:
\begin{itemize}
    \item \textbf{\textit{Inefficient 2D-3D Projection.}} Considering the domain gap between 2D and 3D, most existing works represent 3D data as its 2D counterpart as input for SAM, and back-project the segmentation results into 3D space, e.g., using additional RGB images~\citep{yang2023sam3d, yin2024sai3d, xu2023sampro3d}, multi-view renderings~\citep{zhou2023partslip++}, or Neural Radiance Field~\citep{cen2023segment}. Such modality transition introduces significant processing complexity, hindering efficient implementation.\vspace{0.05cm}

    \item \textbf{\textit{Degradation of 3D Spatial Information.}} The reliance on 2D projections results in the loss of fine-grained 3D geometries and semantics, as multi-view data often fails to preserve spatial relations. Furthermore, the internal structures of 3D objects cannot be adequately captured by 2D images, significantly limiting segmentation accuracy.\vspace{0.05cm}

    \item \textbf{\textit{Loss of Prompting Flexibility.}} A compelling strength of SAM lies in its interactive capabilities through various prompt alternatives. Unfortunately, these functionalities are mostly disregarded in current methods, as users struggle to specify precise 3D positions using 2D representations. Consequently, SAM is typically used for dense segmentation across entire multi-view images, thereby sacrificing interactivity.\vspace{0.05cm}

    \item \textbf{\textit{Limited Domain Transferability.}} Existing 2D-3D projection techniques are often tailored to specific 3D scenarios, heavily dependent on in-domain patterns. This makes them challenging to apply to new contexts, e.g., from objects to scenes or from indoor to outdoor environments. 
    Another research direction~\citep{zhou2024point} aims to train a promptable network from scratch in 3D. While bypassing the need for 2D projections, it demands substantial training and data resources and may still be constrained by training data distributions.\vspace{0.05cm}

\end{itemize}

\begin{table*}[t]
\centering
\small
\caption{\textbf{Comparison of \method and Previous SAM-based Methods}~\citep{yang2023sam3d,cen2023segment,xu2023sampro3d,zhou2024point}. To our best knowledge, \method presents the most faithful implementation of SAM~\citep{kirillov2023segment} in 3D, demonstrating superior implementation efficiency, promptable flexibility, and generalization capabilities for 3D segmentation.}
\vspace{0.3cm}
\begin{adjustbox}{width=\textwidth}
	\begin{tabular}{lccccccccc}
	\toprule
	  \multirow{2}*{\makecell[c]{\vspace{-0.2cm}Method}} &\multirow{2}*{\makecell[c]{\vspace{-0.2cm}Zero-shot}} &\multirow{2}*{\makecell[c]{\vspace{-0.2cm}Project-free}} &\multicolumn{3}{c}{3D Prompt} &\multicolumn{4}{c}{3D Scenario} \\
   \cmidrule(lr){4-6} \cmidrule(lr){7-10}
   & & &\makecell[c]{Point} &\makecell[c]{Box} &\makecell[c]{Mask} &\makecell[c]{Object} &\makecell[c]{Indoor} &\makecell[c]{Outdoor} &\makecell[c]{Raw LiDAR}\\
   \cmidrule(lr){1-1} \cmidrule(lr){2-10}
        SAM3D &\checkmark &- &- &- &- &- &\checkmark &- &-\\
        SA3D &\checkmark &- &- &- &- &- &\checkmark &\checkmark &-\\
        SAMPro3D &\checkmark &- &- &- &- &- &\checkmark &- &-\\
        Point-SAM &- &\checkmark &\checkmark &- &- &\checkmark &\checkmark &\checkmark &- \\
        
        \cmidrule(lr){1-10}
        \rowcolor{skyblue!15}
        \makecell[c]{\textbf{\method}} &{\checkmark} &{\checkmark} &{\checkmark} &{\checkmark} &{\checkmark} &{\checkmark} &{\checkmark} &{\checkmark} &{\checkmark}\\
	\bottomrule
	\end{tabular}
\end{adjustbox}
\label{t1}
\end{table*}%

In this project, we introduce \method, adapting SAM 2 for efficient, projection-free, promptable, and zero-shot 3D segmentation.
As an initial step in this direction, our target is not to push the performance limit, but rather to demonstrate the potential of SAM in achieving robust and effective 3D segmentation in diverse contexts.
Specifically, \method exhibits three features as outlined:

\begin{itemize}

\item \textbf{\textit{Segmenting Any 3D as Videos.}}
To preserve 3D geometries during segmentation, while ensuring compatibility with SAM 2, we adopt voxelization to mimic a video. Voxelized 3D data, with a shape of $w \times h \times l \times 3$, closely resembles the format of videos of $w \times h \times t \times 3$. This representation allows SAM 2 for zero-shot 3D segmentation while retaining sufficient spatial information, without the need of additional training or 2D-3D projection.\vspace{0.05cm}

\item \textbf{\textit{Supporting Multiple 3D Prompts.}}
Building on SAM 2, \method supports three types of prompts: 3D points, bounding boxes, and masks. Starting with a user-provided 3D prompt, e.g., a point (x, y, z), we divide the 3D space into three orthogonal directions, generating six corresponding videos. Then, the multi-directional segmentation results are integrated to form the final prediction in 3D space, allowing for interactive promptable segmentation.\vspace{0.05cm}

\item \textbf{\textit{Generalizable to Various Scenarios.}}
With our concise framework, \method demonstrates strong generalization capabilities in diverse 3D scenarios with varying point cloud distributions. As showcased in Figure~\ref{f1}, our approach can effectively segment single objects, indoor scenes, outdoor scenes, and raw LiDAR, highlighting its superior transferability across different domains.\vspace{0.05cm}

\end{itemize}

\section{\method}
\label{s2}

\begin{figure*}[t]
  \centering
\includegraphics[width=\textwidth]{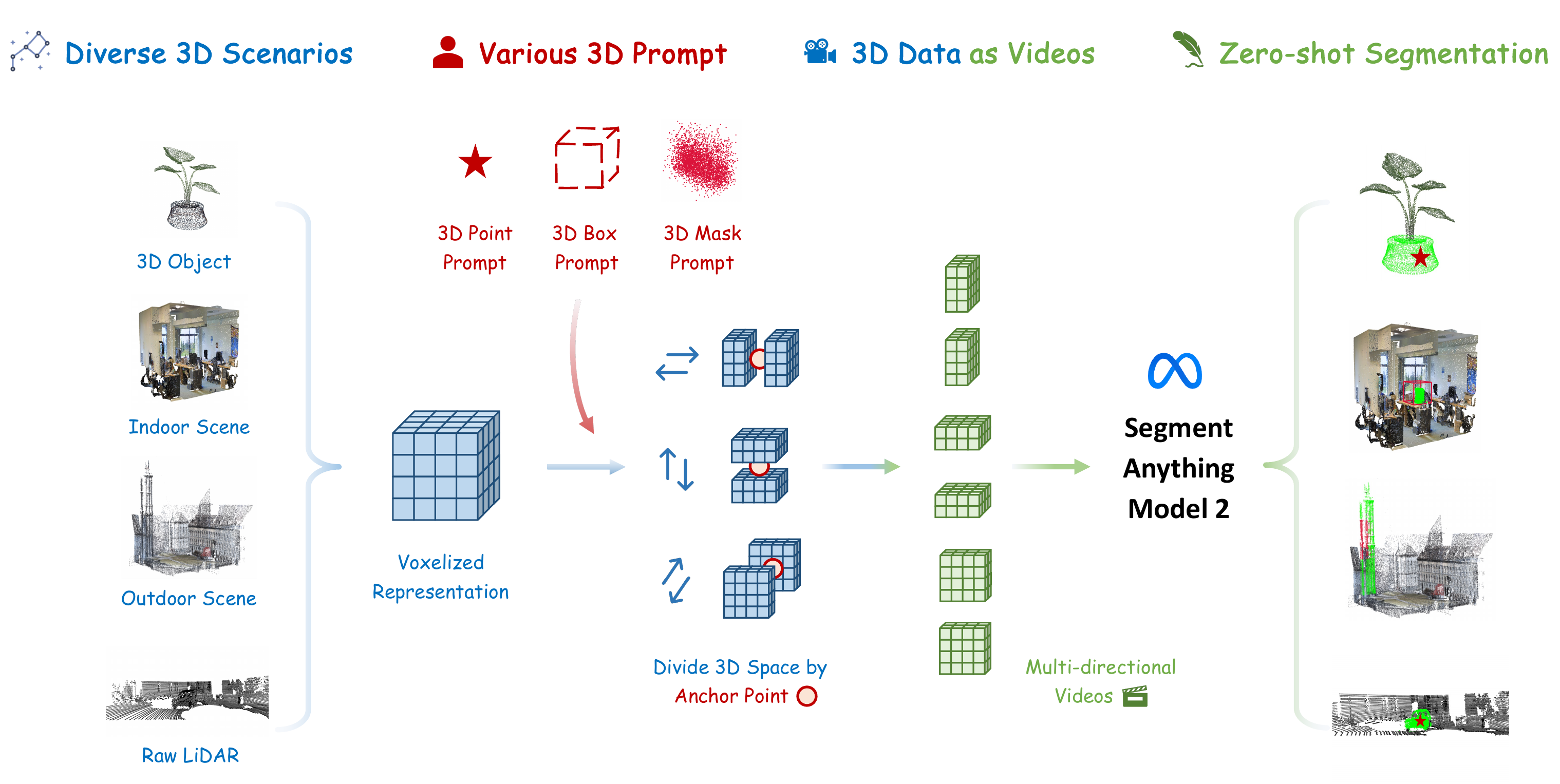}
   \caption{\textbf{The Detailed Methodology of \method.} We convert any input 3D data into voxelized representations, and utilize user-provided 3D prompt to divide the 3D space along six directions, effectively simulating six different videos for SAM 2 to perform zero-shot segmentation.}
    \label{f2}
\end{figure*}

The detailed methodology of \method is presented in Figure~\ref{f2}. In Section~\ref{s2.1}, we introduce how \method efficiently formats 3D data for compatibility with SAM 2~\citep{ravi2024sam}, avoiding complex projection process. Then, in Section~\ref{s2.2}, we detail the three types of 3D prompt supported and their associated segmentation techniques. Finally, in Section~\ref{s2.3}, we illustrate four challenging 3D scenarios effectively addressed by \method.

\subsection{3D Data as Videos}
\label{s2.1}

Given any object-level or scene-level point cloud, we denote it by $P \in \mathbb{R}^{n\times 6}$, with each point as $p = (x, y, z, r, g, b)$. Our aim is to convert $P$ into a data format that, for one hand, SAM 2 can directly process in a zero-shot manner, and, for the other, the fine-grained spatial geometries can be well preserved. To this end, we adopt the 3D voxelization technique. Compared to RGB image mapping~\citep{yang2023sam3d, yin2024sai3d, xu2023sampro3d}, multi-view rendering~\citep{zhou2023partslip++}, and NeRF~\citep{cen2023segment} in previous efforts, voxelization is efficiently performed in 3D space, thereby free from information degradation and cumbersome post-processing.

In this way, we obtain a voxelized representation of the 3D input, denoted by $V \in \mathbb{R}^{w\times h\times l\times 3}$ with each voxel as $v = (r, g, b)$. For simplicity, the $(r, g, b)$ value is set according to the point nearest to the voxel center. This format closely resembles videos with a shape of $w\times h\times t\times 3$. The main difference is that, video data contains unidirectional temporal dependency across $t$ frames, while 3D voxels are isotropic along three spatial dimensions. Considering this, we convert the voxel representation as a series of multi-directional videos, inspiring SAM 2 to segment 3D the same way as videos.

\subsection{Promptable Segmentation}
\label{s2.2}

For flexible interactivity, our \method supports three types of prompt in 3D space, which can be utilized either separately or jointly. We specify the prompting and segmentation details below:

\begin{itemize}
    \item \textbf{3D Point Prompt}, denoted as $p_p = (x_p, y_p, z_p)$. We first regard $p_p$ as an anchor point in 3D space to define three orthogonal 2D sections. Starting from these sections, we divide the 3D voxels into six subparts along six spatial directions, i.e., front, back, left, right, up, and down. Then, we regard them as six different videos, where the section serves as the first frame and $p_p$ is projected as the 2D point prompt. After applying SAM 2 for concurrent segmentation, we integrate the results of six videos as the final 3D mask prediction.\vspace{0.05cm}
    
    \item \textbf{3D Box Prompt}, denoted as $b_p = (x_p, y_p, z_p, w_p, h_p, l_p)$, including 3D center coordinates and dimensions. We adopt the geometric center of $b_p$ as the anchor point, and represent the 3D voxels by six different videos as aforementioned. For video of a certain direction, we project $b_p$ into the corresponding 2D section to serve as the box point for segmentation. We also support 3D box with rotation angles, e.g., ($\alpha_p$, $\beta_p$, $\gamma_p$), for which the bounding rectangle of projected $b_p$ is adopted as the 2D prompt.\vspace{0.05cm}
    
    \item \textbf{3D Mask Prompt}, denoted as $M_p \in \mathbb{R}^{n\times 1}$, where 1 or 0 indicates the masked and unmasked areas. We employ the center of gravity of the mask prompt as the anchor point, and divide 3D space into six videos likewise. The intersection between the 3D mask prompt and each section is utilized as the 2D mask prompt for segmentation. This type of prompting can also serve as a post-refinement step to enhance the accuracy of previously predicted 3D masks.\vspace{0.05cm}

\end{itemize}

\subsection{Any 3D Scenarios}
\label{s2.3}

With our concise framework design, \method exhibits superior zero-shot generalization performance across diverse domains, ranging from objects to scenes and indoor to outdoor environments. We elaborate on four distinct 3D scenarios below:

\begin{itemize}
    \item \textbf{3D Object}, e.g., Objaverse~\citep{deitke2023objaverse}, with a wide array of categories, possesses unique characteristics across different instances, including colors, shapes, and geometries. Adjacent components of an object might overlap, occlude, or integrate with each other, which requires models to accurately discern subtle differences for part segmentation.\vspace{0.05cm}
    
    \item \textbf{Indoor Scene}, e.g., S3DIS~\citep{armeni20163d} and ScanNet~\citep{dai2017scannet}, are typically characterized by multiple objects arranged within confined spaces, like rooms. The complex spatial layouts, similarity in appearance, and varied orientations between objects pose challenges for models to segment them from backgrounds.\vspace{0.05cm}
    
    \item \textbf{Outdoor Scene}, e.g., Semantic3D~\citep{hackel2017semantic3d}, differs from indoor scenes, primarily due to the stark size contrasts of objects (buildings, vehicles, and humans) and the larger scale of point clouds (from a room to an entire street). These variations complicates the segmentation of objects whether at a global scale or a fine-grained level.\vspace{0.05cm}
    
    \item \textbf{Raw LiDAR}, e.g., KITTI~\citep{Geiger2012CVPR} in autonomous driving, is distinct from typical point clouds for its sparse distribution and absence of RGB information.
    The sparsity demands models to infer missing semantics for understanding the scene, and the lack of colors enforces models to only rely on geometric cues to differentiate between objects. In \method, we directly set the RGB values of 3D voxels by the LiDAR intensity.\vspace{0.05cm}

\end{itemize}

\section{Discussion and Insight}
\label{s2.5}

Building on the effectiveness of \method, we delve into two compelling yet challenging issues within the realm of 3D, and share our insights on future multi-modality learning.

\subsection{How to Adapt 2D Foundation Models to 3D?}

The availability of large-scale, high-quality data has significantly empowered the development of large models in language~\citep{gpt3,llama,zhang2023llama}, 2D vision~\citep{llava,gemini,internvl}, and vision-language~\citep{gao2024sphinx,liu2023improvedllava,li2024llava,zhang2024mavis} domains. In contrast, the 3D field has long struggled with a scarcity of data, hindering the training of large 3D models. As a result, researchers have turned to the alternative of transferring pre-trained 2D models into 3D.

\textit{The primary challenge lies in bridging the modal gap between 2D and 3D.} Pioneering approaches, such as PointCLIP~\citep{zhang2022pointclip}, its V2~\citep{Zhu2022PointCLIPV2}, and subsequent methods~\citep{ji2023jm3d,huang2023clip2point}, project 3D data into multi-view images, which encounter implementation inefficiency and information loss. Another line of work, including ULIP series~\citep{ulip,ulip2}, I2P-MAE~\citep{zhang2023learning}, and others~\citep{liu2023openshape,qi2023recon,guo2023joint}, employs knowledge distillation using 2D-3D paired data. While this method generally performs better due to extensive training, it suffers from limited 3D transferability in out-of-domain scenarios. Recent efforts have also explored more complex and costly solutions, such as joint multi-modal spaces (e.g., Point-Bind \& Point-LLM~\citep{guo2023point}), larger-scale pre-training (Uni3D~\citep{zhou2023uni3d}), and virtual projection techniques (Any2Point~\citep{tang2024any2point}).

From \method, we observe that \textit{representing 3D data as videos through voxelization may offer an optimal solution}, providing a balanced trade-off between performance and efficiency. This approach not only preserves the spatial geometries inherent in 3D space with a simple transformation, but also presents a grid-based data format that 2D models can directly process. Despite this, further experiments are necessary to validate and reinforce this observation.

\subsection{What is the Potential of \method in 3D Domains?}

To the best of our knowledge, \textit{\method presents the most accurate and comprehensive implementation of SAM in 3D}, successfully inheriting its implementation efficiency, promptable flexibility, and generalization capabilities. While previous SAM-based approaches~\citep{yang2023sam3d, xu2023sampro3d,yin2024sai3d} have achieved 3D segmentation, they often fall short in scalability and transferability to benefit other 3D tasks. In contrast, inspired by SAM in 2D domains, \method demonstrates significant potential to advance various 3D applications.

For fundamental 3D understanding, \method can serve as a unified initialized backbone for further fine-tuning, offering strong 3D representations simultaneously across 3D objects, indoor scenes, outdoor scenes, and raw LiDAR. In the context of training large 3D models, \method can be employed as an automatic data annotation tool, which mitigates the data scarcity issue by generating large-scale segmentation labels across diverse scenarios. For 3D and language-vision learning, \method inherently provides a joint embedding space across 2D, 3D, and video domains, due to its zero-shot capabilities, which could further enhance the effectiveness of models like Point-Bind~\citep{guo2023point}. Additionally, in the development of 3D large language models (LLMs)~\citep{3dllm,xu2023pointllm,wang2023chat,guo2023point}, \method can function as a powerful 3D encoder, supplying LLMs with 3D tokens, and leveraging its promptable features to equip LLMs with promptable instruction-following capabilities.

\section{Demos}
\label{s3}

\begin{figure*}[t]
\vspace{-0.4cm}
\centering
\includegraphics[width=\textwidth]{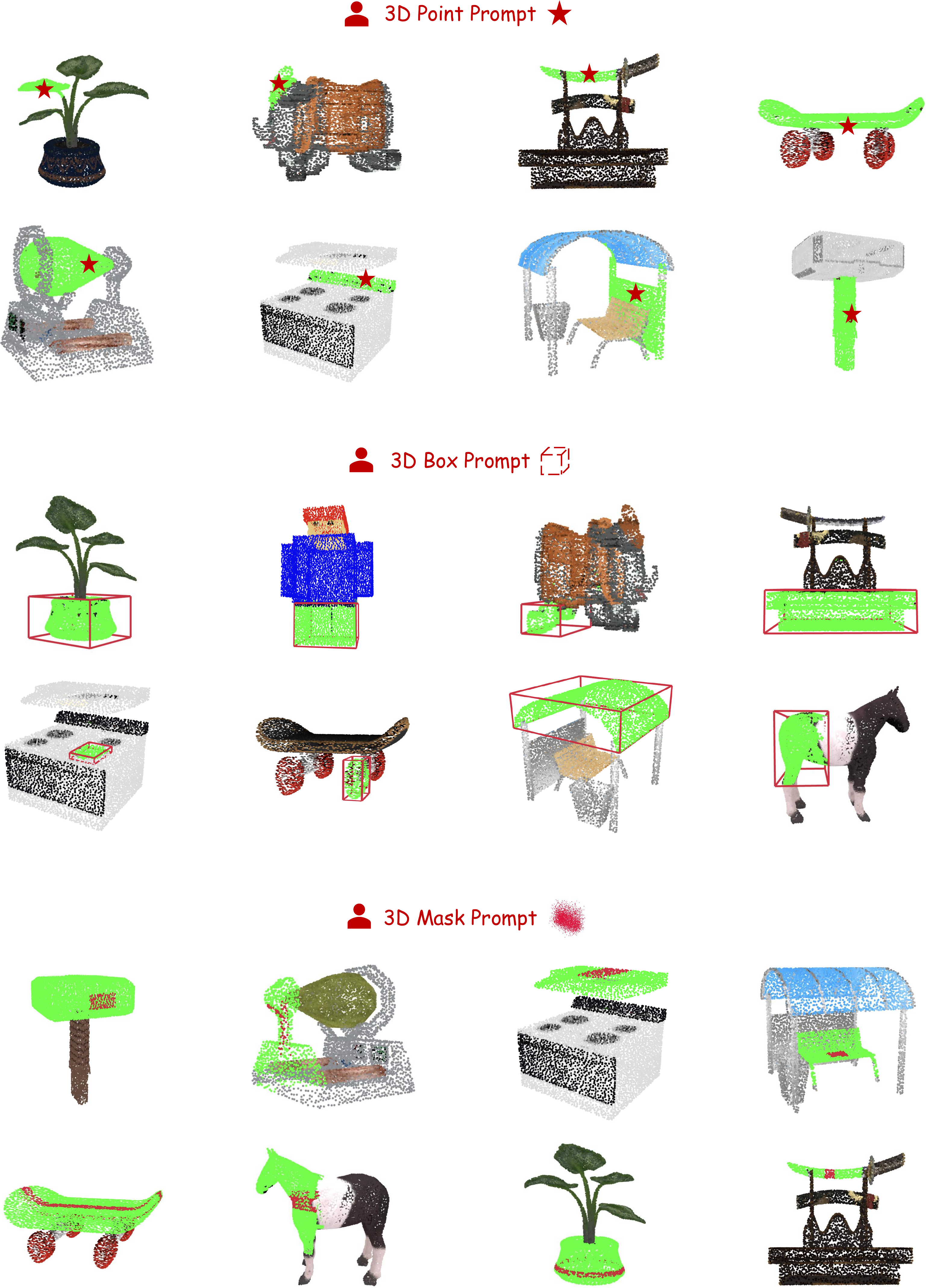}
\vspace{0.1cm}
   \caption{\textbf{3D Object Segmentation with \method on Objaverse~\citep{deitke2023objaverse}.} The \textcolor{red}{3D prompt} and \textcolor{ggreen}{segmentation results} are highlighted in red and green, respectively.}
   \label{object}
\end{figure*}

\begin{figure*}[t]
\centering
\includegraphics[width=\textwidth]{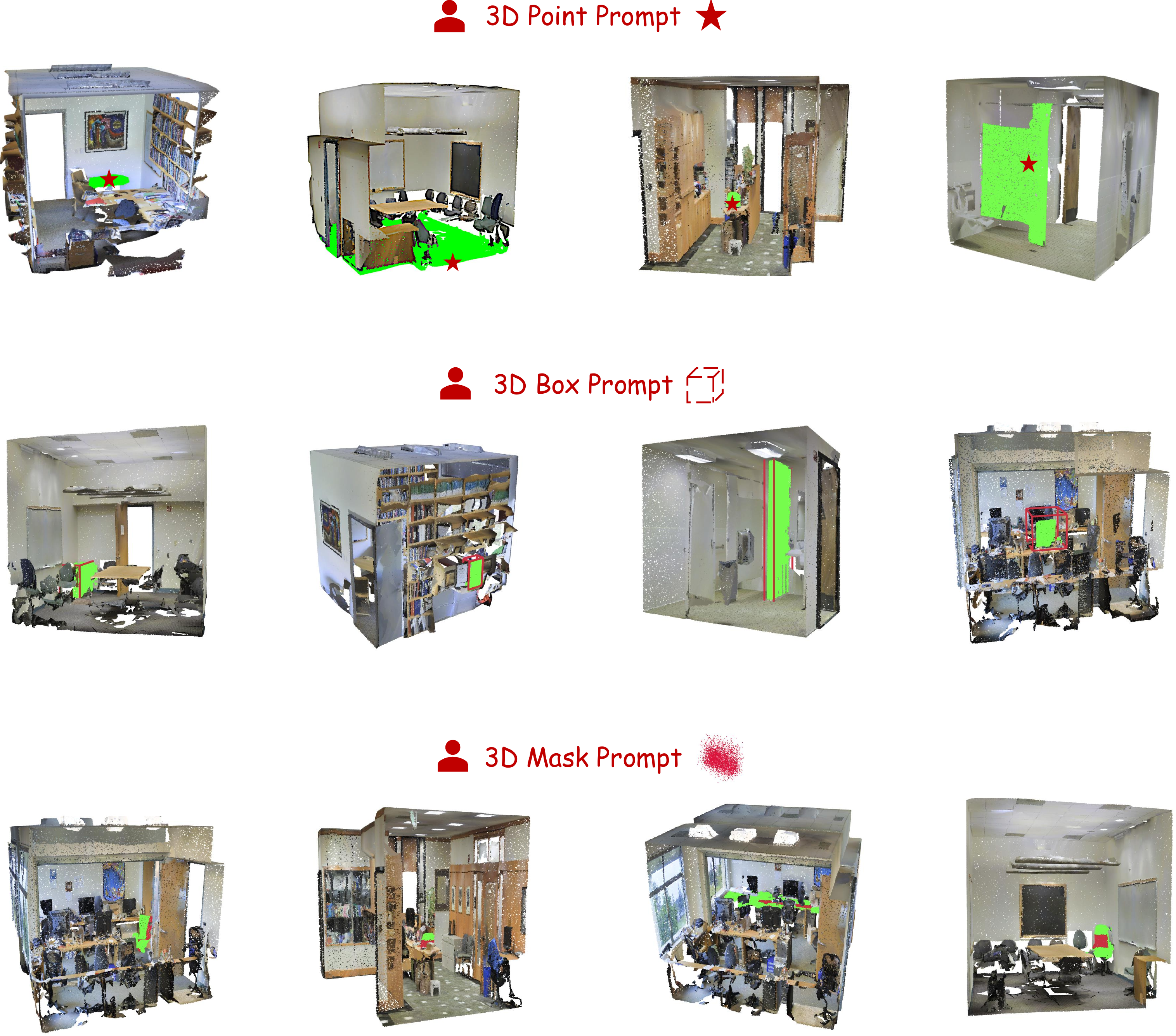}
\vspace{0.1cm}
   \caption{\textbf{3D Indoor Scene Segmentation with \method on S3DIS~\citep{armeni20163d}.} The \textcolor{red}{3D prompt} and \textcolor{ggreen}{segmentation results} are highlighted in red and green, respectively.}
   \label{s3dis}
\end{figure*}

\begin{figure*}[t]
\centering
\includegraphics[width=\textwidth]{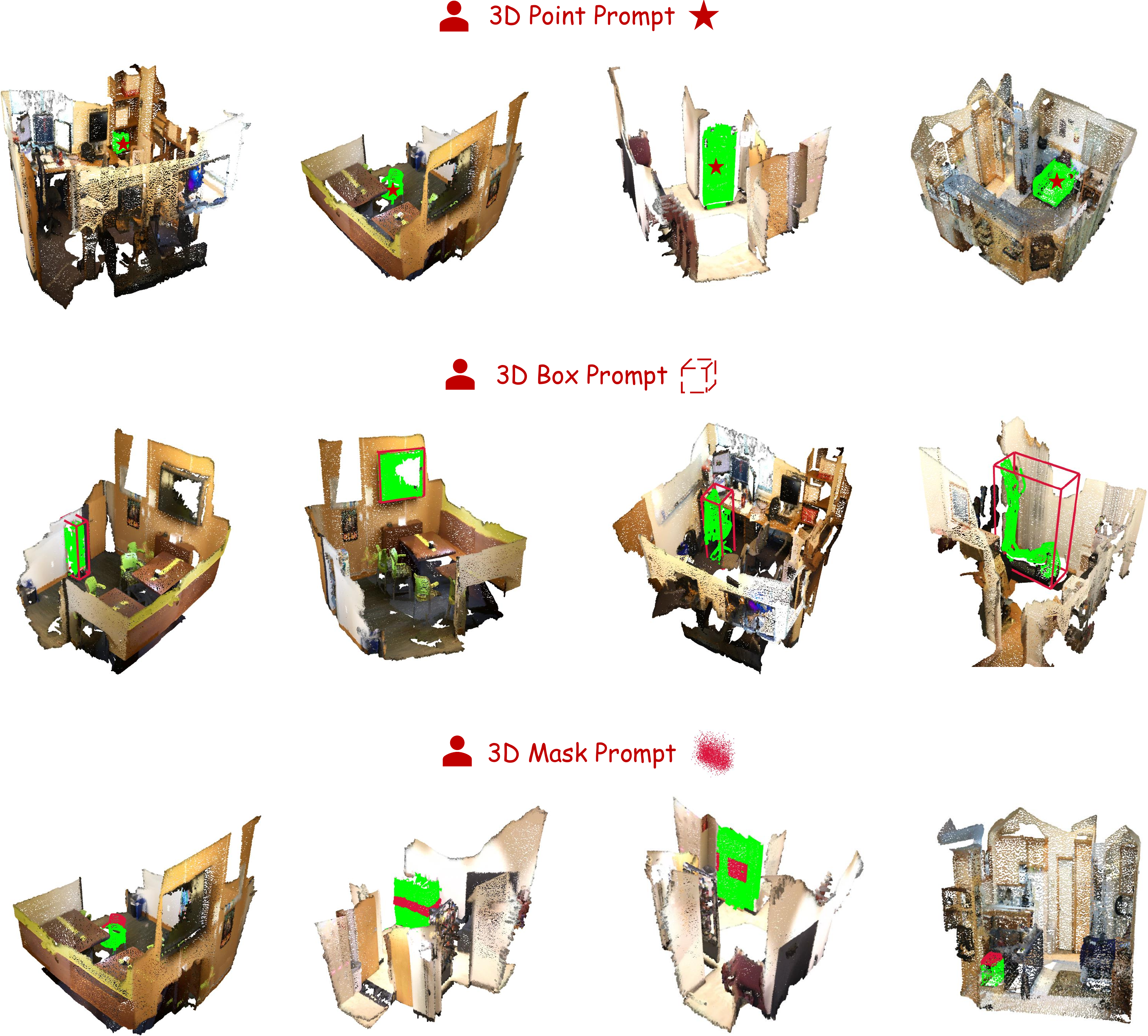}
\vspace{0.1cm}
   \caption{\textbf{3D Indoor Scene Segmentation with \method on ScanNet~\citep{dai2017scannet}.} The \textcolor{red}{3D prompt} and \textcolor{ggreen}{segmentation results} are highlighted in red and green, respectively.}
   \label{scannet}
\end{figure*}

\begin{figure*}[t]
\centering
\includegraphics[width=\textwidth]{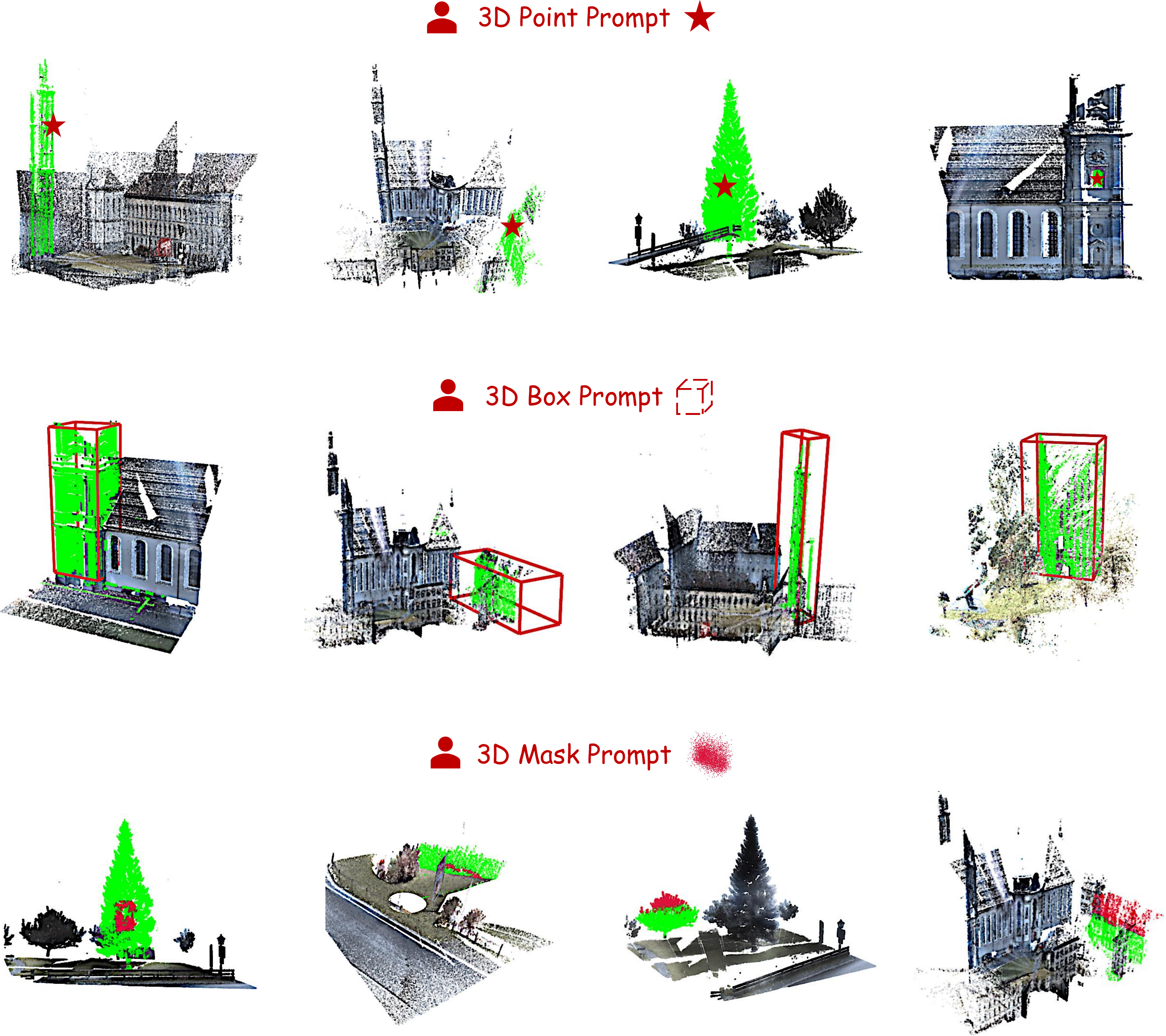}
\vspace{0.1cm}
   \caption{\textbf{3D Outdoor Scene Segmentation with \method on Semantic3D~\citep{hackel2017semantic3d}.} The \textcolor{red}{3D prompt} and \textcolor{ggreen}{segmentation results} are highlighted in red and green, respectively.}
   \label{outdoor}
\end{figure*}

\begin{figure*}[t]
\centering
\includegraphics[width=\textwidth]{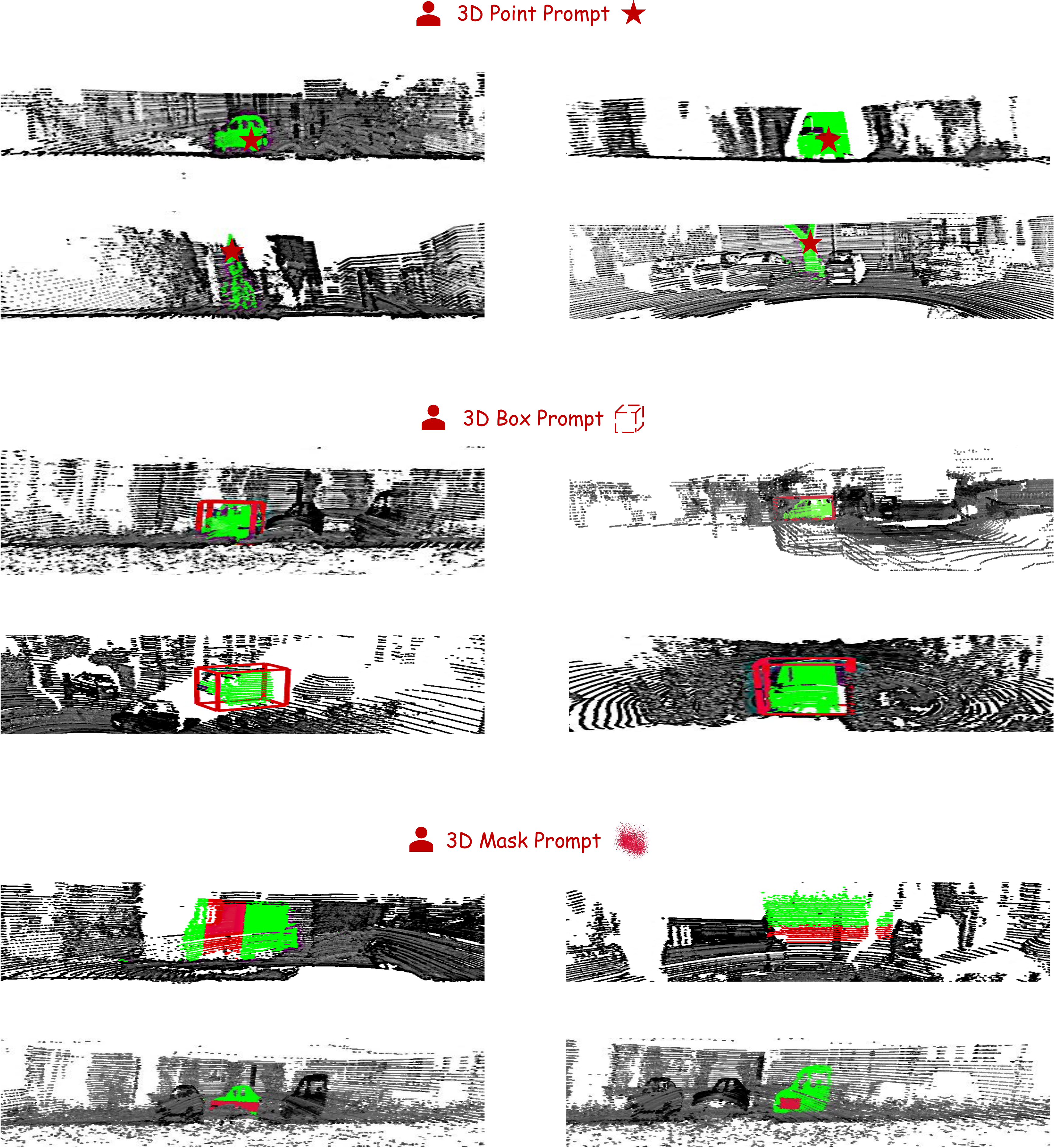}
\vspace{0.1cm}
   \caption{\textbf{3D Raw LiDAR Segmentation with \method on KITTI~\citep{Geiger2012CVPR}.} The \textcolor{red}{3D prompt} and \textcolor{ggreen}{segmentation results} are highlighted in red and green, respectively.}
   \label{lidar}
\end{figure*}

In Figures~\ref{object}-\ref{lidar}, we showcase demonstrations of \method in segmenting 3D data with various 3D prompt on different datasets~\citep{deitke2023objaverse,armeni20163d,dai2017scannet,hackel2017semantic3d,Geiger2012CVPR}. For further implementation details, please refer to our open-sourced code.

\clearpage

\section{Conclusion}
\label{s4}

In this project, we propose \method, which leverages Segment Anything 2 (SAM 2) to 3D segmentation with a zero-shot and promptable framework. By representing 3D data as multi-directional videos, \method supports various types of user-provided prompt (3D point, box, and mask), and exhibits robust generalization across diverse 3D scenarios (3D object, indoor scene, outdoor environment, and raw sparse LiDAR). As a preliminary investigation, \method provides unique insights into adapting SAM 2 for effective and efficient 3D understanding. We hope our method may serve as a foundational baseline for promptable 3D segmentation, encouraging further research to fully harness SAM 2's potential in 3D domains.

\clearpage
\bibliography{iclr2024_conference}
\bibliographystyle{iclr2024_conference}


\end{document}

%% file: math_commands.tex
\usepackage{amsmath,amsfonts,bm}

\def\eqref#1{equation~\ref{#1}}

\def\1{\bm{1}}

\DeclareMathAlphabet{\mathsfit}{\encodingdefault}{\sfdefault}{m}{sl}
\SetMathAlphabet{\mathsfit}{bold}{\encodingdefault}{\sfdefault}{bx}{n}

